\documentclass[10pt,twocolumn,letterpaper]{article}

\usepackage{cvpr}
\usepackage{times}
\usepackage{epsfig}
\usepackage{graphicx}
\usepackage{amsmath}
\usepackage{amssymb}
\graphicspath{ {report_figures/} }

\usepackage[breaklinks=true,bookmarks=false]{hyperref}

\cvprfinalcopy % *** Uncomment this line for the final submission

\def\cvprPaperID{****} % *** Enter the CVPR Paper ID here

\begin{document}

%%%%%%%%% TITLE
\title{Quantifying Translation-Invariance in Convolutional Neural Networks}

\author{Eric Kauderer-Abrams\\
Stanford University\\
{\tt\small ekabrams@stanford.edu}
}

\maketitle
%\thispagestyle{empty}

%%%%%%%%% ABSTRACT
\begin{abstract}
A fundamental problem in object recognition is the development of image representations that are invariant to common transformations such as translation, rotation, and small deformations. There are multiple hypotheses regarding the source of translation invariance in CNNs. One idea is that translation invariance is due to the increasing receptive field size of neurons in successive convolution layers. Another possibility is that invariance is due to the pooling operation. We develop a simple a tool, the translation-sensitivity map, which we use to visualize and quantify the translation-invariance of various architectures. We obtain the surprising result that architectural choices such as the number of pooling layers and the convolution filter size have only a secondary effect on the translation-invariance of a network. Our analysis identifies training data augmentation as the most important factor in obtaining translation-invariant representations of images using convolutional neural networks. 
\end{abstract}

%%%%%%%%% BODY TEXT
\section{Introduction}

Object recognition is a problem of fundamental importance in visual perception. The ability to extrapolate from raw pixels to the concept of a coherent object persisting through space and time is a crucial link connecting low-level sensory processing with higher-level reasoning. A fundamental problem in object recognition is the development of image representations that are invariant to common transformations such as translation, rotation, and small deformations. Intuitively, these are desirable properties for an object recognition algorithm to have: a picture of a cat should be recognizable regardless of the cat's location and orientation within the image. It has been demonstrated that affine transformations account for a significant portion of intra-class variability in many datasets \cite{soatto1}. 
	
In recent years, several different approaches to the invariance problem have emerged. Loosely modeled on the local receptive fields and hierarchical structure of Visual Cortex, Convolutional Neural Networks (CNNs) have achieved unprecedented accuracy in object recognition tasks \cite{lecun} \cite{hinton}. There is widespread consensus in the literature that CNNs are capable of learning translation-invariant representations \cite{transformer} \cite{deep_sym} \cite{inv_features}. However, there is no definitive explanation for the mechanism by which translation-invariance is obtained.

One idea is that translation-invariance is due to the gradual increase in the receptive field size of neurons in successive convolution layers \cite{deep_sym}. An alternative hypothesis is that the destructive pooling operations are responsible for translation-invariance \cite{transformer}. Despite the widespread consensus, there has been little work done in exploring the nature of translation-invariance in CNNs. In this study we attempt to fill this gap, directly addressing the two questions: To what extent are the representations produced by CNNs translation invariant? Which features of CNNs are responsible for this translation invariance? 

	To address these questions we introduce a simple tool for visualizing and quantifying translation invariance, called translation-sensitivity maps. Using translation-sensitivity maps, we quantify the degree of translation-invariance of several CNN architectures trained on an expanded version of the MNIST digit dataset \cite{mnist}. We obtain the surprising result that the most important factor in obtaining translation-invariant CNNs is training data augmentation and not the particular architectural choices that are often discussed.

%------------------------------------------------------------------------
\section{Related Work}

There have been several previous works addressing the invariance problem in CNNs. A recent study quantified the invariance of representations in CNNs to general affine transformations such as rotations and reflections \cite{meas_inv}. There have also been recent developments in altering the structure of CNNs and adding new modules to CNNs to increase the degree of invariance to common transformations \cite{deep_sym}. A poignant example is spatial transformer networks \cite{transformer}, which include a new layer that learns an example-specific affine transformation to be applied to each image. The intuition behind this approach, that the network can transform any input into a standard form with the relevant information centered and oriented consistently, has been confirmed empirically. 

There have been several studies done on the theory behind the invariance problem. One such example is The Scattering Transform \cite{mallat1} \cite{mallat2}, which computes representations of images that are provably invariant to affine transformations and stable to small deformations. Furthermore, it does so without any learning, relying instead on a cascade of fixed wavelet transforms and nonlinearities. With performance matching CNNs on many datasets, The Scattering Transform provides a powerful theoretical foundation for addressing the invariance problem. In addition there are several related works addressing the invariance problem for scenes rather than images \cite{soatto2} and in a more general unsupervised learning framework \cite{poggio}. 

	The methods that we describe in the next section can be applied to any of the above approaches to quantify the degree of translation-invariance achieved by the algorithms. 

%-------------------------------------------------------------------------

\section{Methods}
\subsection{Convolutional Neural Networks}
This section describes the development of tools used to quantify the translation-invariance of CNNs, and the setup for the experiments to be run using these tools. It is important to point out that we refer to a network as ``translation-invariant'' as shorthand for saying that the network output varies little when the input is translated. In practice, no algorithm is truly translation-invariant; instead, we measure the sensitivity of the output to translation in the input. We loosely call an architecture more translation-invariant than another if its output is less sensitive to translations of the input. Now we introduce tools to make comparisons between architectures more precise. 

	Our approach entails constructing and training many different CNN architectures. To provide some background: a CNN is a hierarchy of several different types of transformations referred to as ``layers'' that take an image as input and produce an output ``score vector'' giving the network's prediction for the category that the image belongs to. Each layer implements a specific mathematical operation. Convolution layers compute the convolution of a small, local filter with the input. A unit in a convolution layer acts as a feature detector, becoming active when the template stored in the filter matches with a region of the input. Convolution layers are followed by the application of a point-wise nonlinearity, which is commonly chosen to be the rectified linear function (ReLu). Many architectures include Fully-Connected (FC) Layers at the end of the hierarchy, which multiply the input (reshaped to be a vector) with a matrix of weights corresponding to the connection ``strength'' between each FC unit and each input vector element. Additional common layers are Pooling Layers, which subsample the input and batch normalization layers, which fix the mean and variance of the output of a given layer.

\begin{figure}[t]
\begin{center}
   \includegraphics[width = 1.0\linewidth]{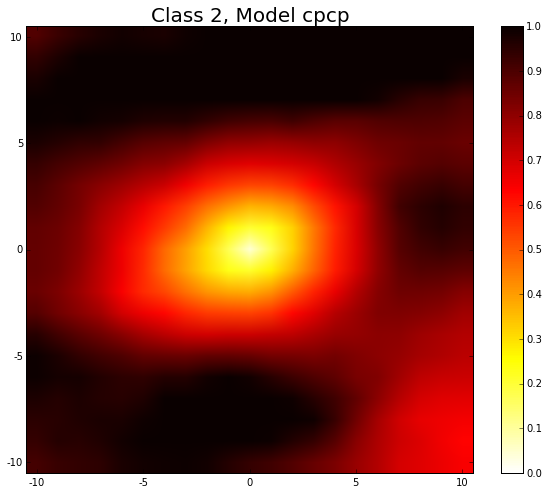}
\end{center}
   \caption{A translation-sensitivity map obtained from a network with two convolution and two pooling layers.}
\end{figure}

To train CNNs, the output of the last layer is fed into a loss function that provides a single scalar value corresponding to the desirability of the network's prediction for a given example. The networks are then trained with backpropogation: computing the gradient of the loss function with respect to all of the parameters in the network. 

For this work we trained several variants of common, relatively small CNNs using the Adam \cite{adam} optimization algorithm and the cross-entropy loss function, $L_{i} = -log(\frac{e^{f_{y_{i}}}}{\sum_{j}e^{f_{j}}})$, with L2 regularization. The learning-rate and regularization strength were selected via the traditional training-validation set approach. All networks were implemented using the TensorFlow framework \cite{tensorflow} on an Intel i7 CPU. 

Given a fully trained CNN, we select an image from the test set to designate as the ``base'' image. Next, we generate all possible translations of the base image that leave the digit fully contained within the frame. For the MNIST dataset centered within a 40x40 pixel frame, the translations went from -10 to 10 pixels in the x-direction and -10 to 10 pixels in the y-direction, where negative and positive are used to denote translation to the left and right in the x-direction, and up and down and in the y-direction. Next, we perform a forward pass on the base image and save the output, which we refer to as the base output vector. We then perform a forward pass on the translated image and compute the Euclidean distance between the base output vector and the translated output vector. Note that for a fully translation-invariant network, this distance would be zero. 

Finally, we divide this distance by the median inter-class distance between all score vectors in a large sample of the data in order to normalize the results so that they can be compared across different networks. This normalization also provides easily interpretable numbers: a distance of one means that the network considers the translated image to be so different from the base image that it might as well belong to another class. A distance of zero corresponds to the case of perfect translation-invariance.

\subsection{Translation-Sensitivity Maps}
	To produce a translation-sensitivity map, we compute the normalized score-space distances as described above between the base image, $I_{0, 0}$, and each translated copy $I_{k_{x}, k_{y}}$, where $I_{k_{x}, k_{y}}$ is the original image translated $k_{x}$ units in the x-direction and $k_{y}$ units in the y-direction. We display the results in a single heat-map in which the color of cell $[k_{x}, k_{y}]$ corresponds to $ d(I_{0, 0}, I_{k_{x}, k_{y}})$. The plots are centered such that the untranslated image is represented in the center of the image, and the colors are set such that whiter colors correspond to lower distances. To produce a single translation-sensitivity map for each input class, we produce separate translation-sensitivity maps using numerous examples of each class from the test set and average the results. The averaging makes sense because the translation-sensitivity maps for different images from a given class are extremely similar. 

\subsection{Radial Translation-Sensitivity Fuctions}
	While the translation-sensitivity maps provide clearly interpretable images quantifying and displaying the nature of translation-invariance for a given trained network, it is unwieldy for quantitatively comparing results between different network architectures. To perform quantitative comparisons, we compute a one-dimensional signature from the translation-sensitivity maps, called a radial translation-sensitivity function, which displays the average value of a translation-sensitivity map at a given radial distance. Intuitively, the radial translation-sensitivity function displays the magnitude of the change in the output of the network as a function of the size of the translation (ignoring the direction). Because most translation-sensitivity maps were anisotropic, the radial translation-sensitivity function provides a rough metric, however it is useful for comparing results across architectures that can be displayed on a single plot.  

	Using the translation-sensitivity maps and radial translation-sensitivity functions, we quantify the translation-invariance of the outputs produced by various CNN architectures with different numbers of convolution and pooling layers. For each network, we train two versions of the network: One trained on the original centered MNIST dataset, and the other trained on an augmented version of the MNIST dataset obtained by adding four randomly-translated copies of each training set image. 

\begin{figure}
\begin{center}
   \includegraphics[width = 0.4\linewidth]{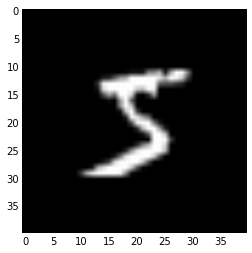}
   \includegraphics[width = 0.4\linewidth]{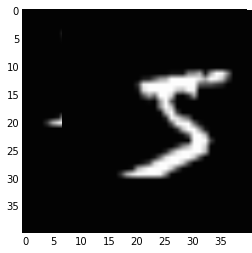}
\end{center}
	 \caption{Untranslated and translated versions of an MNIST image}
\end{figure}

\subsection{Translated MNIST Dataset}
For all experiments we used the MNIST digit dataset. The original images were centered within a 40x40 black frame in order to provide ample space for translating the images. The un-augmented training dataset consists of the 50,000 original centered MNIST images. To generate the augmented dataset, we randomly selected 12,500 of the original centered images and produced 4 randomly translated versions of each of these images to produce to the augmented dataset. The translations in the x and y directions were each chosen randomly from a uniform distribution ranging from one to ten pixels. It is important that the augmented and original training sets are the same size in order to separate the effects of augmentation from the effects of simply having more training data. The dataset also includes 10,000 validation set and 10,000 test set images. 
%-------------------------------------------------------------------------

\begin{figure*}
\begin{center}
\includegraphics[width = 1.0\linewidth]{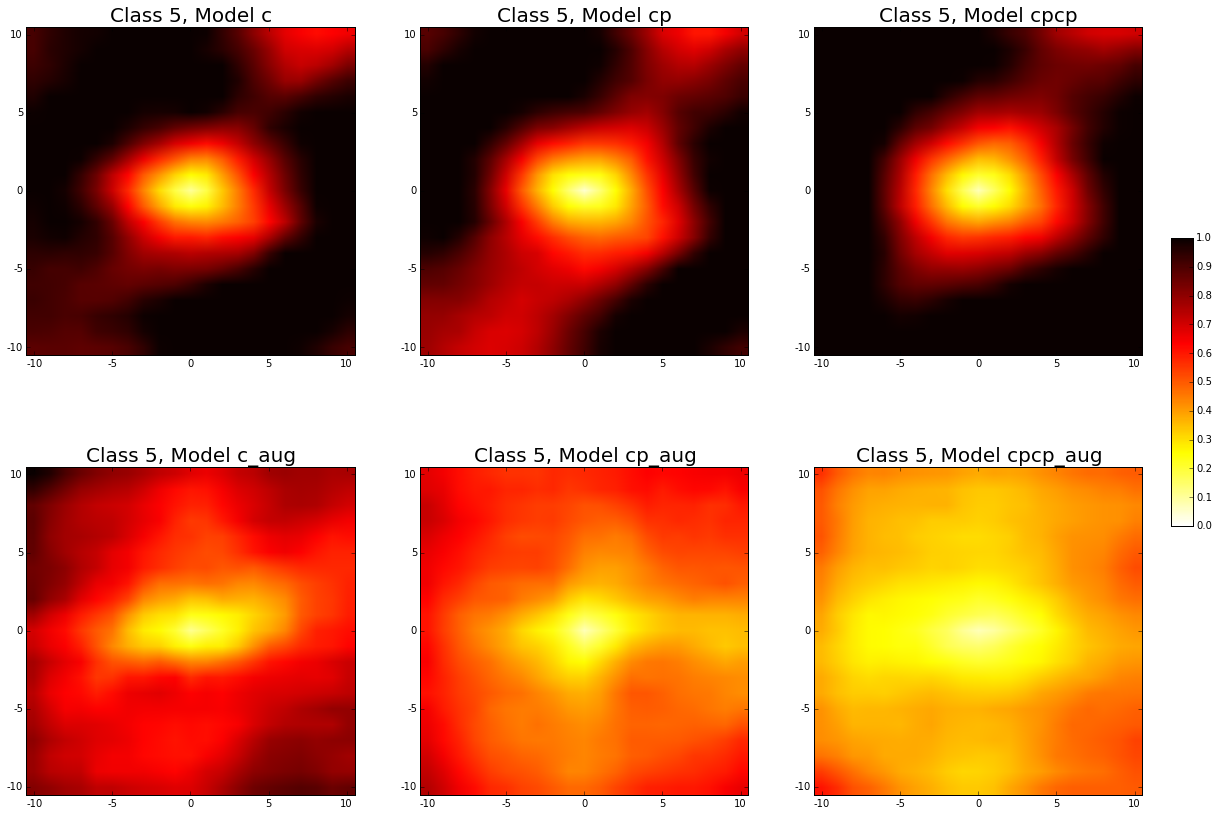}
\end{center}
   \caption{Translation-Sensitivity Maps for three different architectures. The images on the top row were obtained from networks trained with non-augmented data. The images on the bottom row were obtained from the same networks trained with augmented data. The increased brightness of the images on bottom row demonstrates the impact of training data augmentation on the translation-invariance of a network.}
\label{fig:short}
\end{figure*}

\section{Results}
\subsection{Experiment One: Five Architectures With and Without Training Data Augmentation}
We begin with a series of experiments in which we use translation-sensitivity maps and radial translation-sensitivity functions to quantify and compare the translation-invariance of several different CNN architectures, each trained both with and without training data augmentation. The goal is to quantify the role played by training data augmentation, the number of convolution layers, the number of max-pooling layers, and the role played by the ordering of the convolution and pooling layers in obtaining translation-invariant CNNs. 

For referring to each network, we use the following naming convention: Each architecture is described by a string in which `c' stands for a convolution layer, and `p' stands for a max-pooling layer. The Ordering of the letters gives the order of the layers in the network starting at the input. The networks trained with data augmentation end with the string `aug'. Unless otherwise stated, all convolution layers use a filter-size of five pixels, stride one, and padding to preserve the input size, while all max-pooling layers use a two-by-two kernel with stride two.

For this experiment, we train five networks in total: with one convolution and no pooling layers, one convolution and one pooling layer, two convolution layers and no pooling layers, and two different networks with the two convolution and two pooling layers in different orders. All of the above networks have ReLu layers after each convolution layer and have a fully connected layer with 100 hidden units. All networks were trained to convergence, achieving test time accuracy of between 98\% and 99\%.

\begin{figure}
\begin{center}
   \includegraphics[width = 1.0\linewidth]{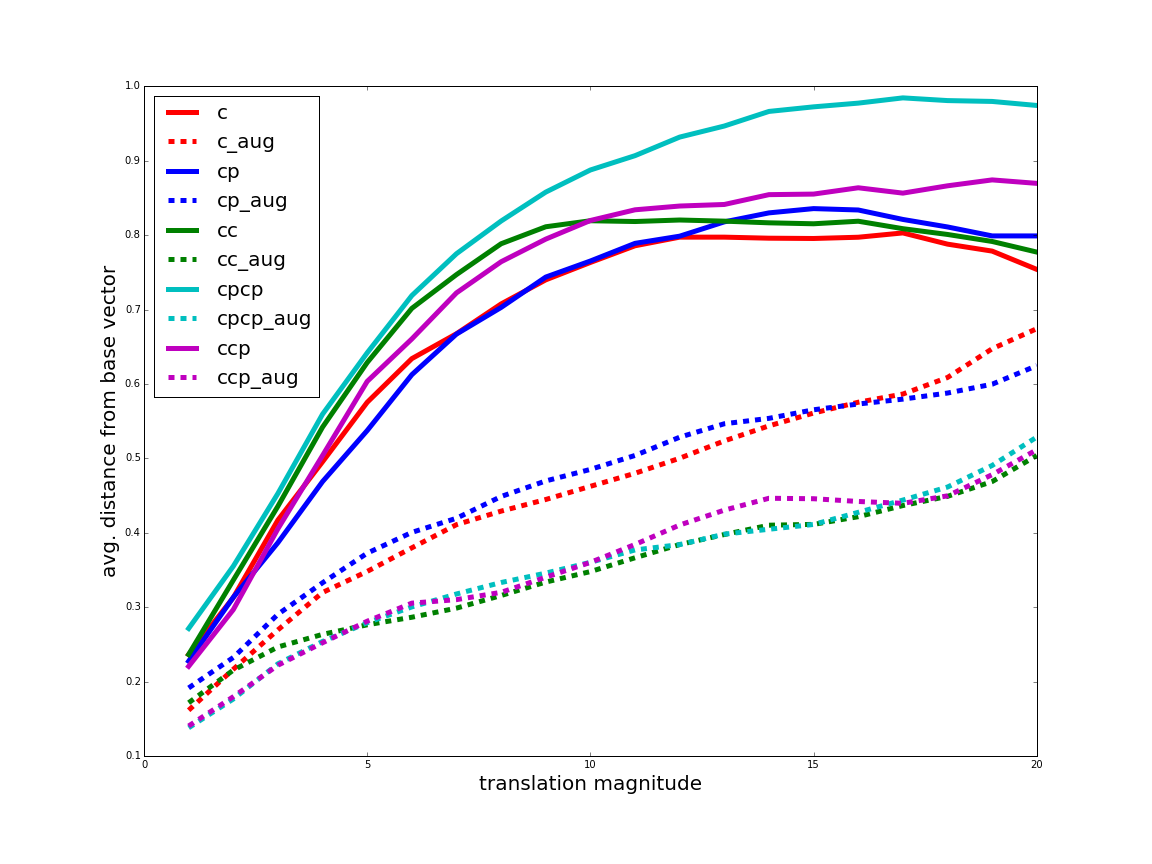}
\end{center}
   \caption{Radial translation-sensitivity function comparing the translation-invariance of five different architectures. The dashed lines are networks trained with augmented data; the solid lines are networks trained with non-augmented data.}
\end{figure}

\subsection{The Importance of Data Augmentation}
The results, a radial translation-sensitivity plot displayed in Figure 4, are partially surprising. The most obvious take-away is that all networks trained on augmented data are significantly more translation-invariant than all networks trained on non-augmented data. From Figures 3 and 4, we see that a simple network with one convolution layer and no pooling layers that is trained on augmented data is significantly more translation-invariant than a network with two convolution and two pooling layers trained on non-augmented data. These results suggest that training data augmentation is the most important factor in obtaining translation-invariant networks. 

\subsection{Architectural Features Play A Secondary Role}
Next, we observe that the architectural choices such as the number of convolution and pooling layers plays a secondary role in determining the translation-invariance of a network. For the networks trained on non-augmented data, there is no correlation between the number of convolution and pooling layers and the translation-invariance of the network. For the networks trained on augmented data, the three network configurations with two convolution and two pooling layers are more translation-invariant than the shallower networks. However, between the three deeper networks there is no distinction, suggesting that it is the network depth, rather than the specific type of layer used that contributes to the translation-invariance of the network. 

\subsection{Experiment Two: The Effect of Filter-Size in a Deeper Network}
In the next experiment, we examine the role of the filter size of the convolution layers in determining the translation-invariance of the network. For this work, we train a deeper network consisting of conv-conv-pool-conv-conv-pool. We chose to use a deeper network because with four convolution layers, the effect of filter size should be more evident than in shallower networks. We trained four instantiations of this architecture: with a filter size of three pixels and filter size of five pixels, and each of these trained on both augmented and non-augmented data. 

The results, displayed in Figure 5 are in line with the results from the previous section. Again, the biggest difference is between the networks trained on augmented data and the networks trained on non-augmented data. Filter size plays a secondary role, in the manner often discussed in the literature, as the network with a larger filter size is more translation invariant than the network with a smaller filter size. 

\begin{figure}[t]
\begin{center}
   \includegraphics[width = 1.0\linewidth]{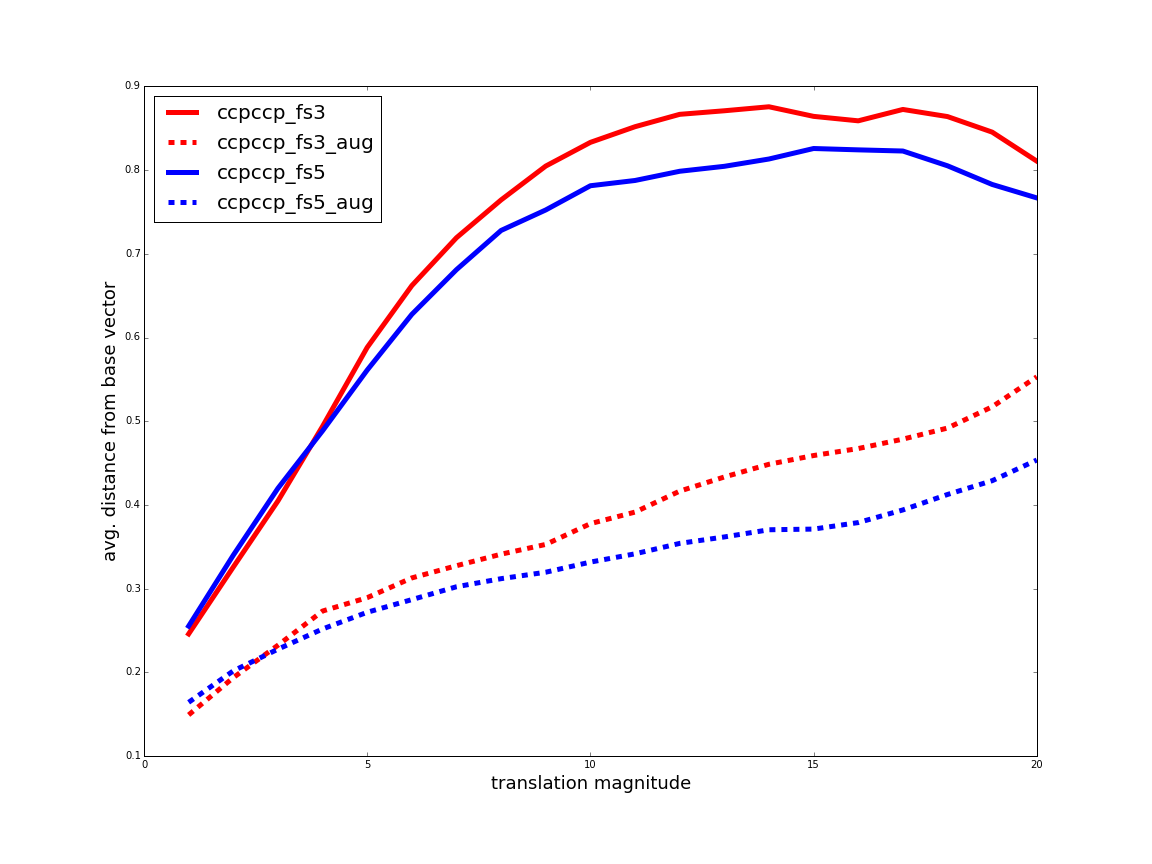}
\end{center}
   \caption{Radial translation-sensitivity function comparing the translation-invariance of the same architecture with  filter sizes of three and five pixels. The dashed lines are networks trained with augmented data; the solid lines are networks trained with non-augmented data. Once again, we see that training data augmentation has the largest effect on the translation-invariance of the network.}
\end{figure}

\section{Conclusion}
In this paper we introduced two simple tools, translation-sensitivity maps and radial translation-sensitivity functions, for visualizing and quantifying the translation-invariance of classification algorithms. We applied these tools to the problem of measuring the translation-invariance of various CNN architectures and isolating the network and training features primarily responsible for the translation-invariance. The results provide a refinement of many ideas circulating around the literature that have not been rigorously tested before this work. 

CNNs are not inherently translation-invariant by virtue of their architecture alone, however they have the capacity to learn translation-invariant representations if trained on appropriate data. The single most important factor in obtaining translation-invariant networks is to train on data that features significant amounts of variation due to translation. Beyond training data augmentation, architectural features such as the number of convolution, the number of pooling layers, and the filter size play a secondary role. Deeper networks with larger filter sizes have the capacity to learn representations that are more translation-invariant but this capacity is only realized if the network is trained on augmented data. 

To follow-up on this work we plan to use translation-sensitivity maps to quantify the translation-invariance achieved by spatial transformer networks. Many promising results have been obtained using Spatial Transformer Networks and it would be interesting to place these within the context of this paper. 

On a higher-level, we are interested in using the invariance problem as a platform for constructing hybrid models between CNNs and algorithms like The Scattering Transform. Given that invariance to nuisance transformations is a desirable property for many object recognition algorithms to have, it seems inefficient that CNNs must learn this from scratch each time. While some have suggested that invariance is inherently built into CNNs through their architecture, our work shows that this is not the case. In contrast, The Scattering Transform, which features no learning in the image representation stage, has translation-invariance built in regardless of how it is trained. On the hand, the ability of CNNs to learn arbitrary patterns from the training data is one of the factors contributing to their considerable success. However, it will be interesting to consider hybrid algorithms which incorporate the desired invariance structure differently. 

{\small
\bibliographystyle{ieee}
\bibliography{egbib}
}

\end{document}